\documentclass[10pt,journal]{IEEEtran}
\usepackage{amsmath,amssymb,graphicx,booktabs,array,xcolor,tikz}
\usetikzlibrary{arrows.meta,calc,positioning}
\usepackage{url}
\graphicspath{{figures/}}

\newcommand{\MSE}{\operatorname{MSE}_{255}}
\newcommand{\LowNetCleanTable}{%
\begin{table}[t]\centering\scriptsize\setlength{\tabcolsep}{3.5pt}
\caption{Clean test mean per-image PSNR (dB), mean $\pm$ three-fit sample SD. Radial24 is supervised with a single-output pixel loss; LowNet24 uses the connected R1--R3 RHS spectral loss. All fixed R1--12 points are retained. A reported $\pm{<}0.001$ denotes nonzero sample SD below 0.001 dB, not zero uncertainty. SD is not a scene-level confidence interval.}\label{tab:clean}
\begin{tabular}{lccc}\toprule
$R$ & Original177 & Radial24(supervised) & LowNet24(unsupervised) \\
\midrule
1 & $43.322\pm0.001$ & $47.273\pm{<}0.001$ & $47.272\pm{<}0.001$ \\
2 & $61.958\pm0.005$ & $74.123\pm0.008$ & $74.136\pm0.008$ \\
3 & $76.585\pm0.016$ & $100.727\pm0.017$ & $100.741\pm0.017$ \\
4 & $89.406\pm0.035$ & $119.909\pm0.012$ & $119.924\pm0.006$ \\
5 & $100.835\pm0.060$ & $121.375\pm0.008$ & $121.378\pm0.012$ \\
6 & $110.738\pm0.076$ & $121.399\pm0.018$ & $121.403\pm0.008$ \\
7 & $117.646\pm0.054$ & $121.419\pm0.006$ & $121.423\pm0.003$ \\
8 & $120.498\pm0.019$ & $121.415\pm0.001$ & $121.407\pm0.010$ \\
9 & $121.099\pm0.015$ & $121.420\pm0.003$ & $121.425\pm0.010$ \\
10 & $121.191\pm0.005$ & $121.424\pm0.010$ & $121.419\pm0.004$ \\
11 & $121.202\pm0.010$ & $121.408\pm0.017$ & $121.419\pm0.007$ \\
12 & $121.201\pm0.009$ & $121.423\pm0.007$ & $121.420\pm0.011$ \\
\bottomrule\end{tabular}\end{table}%
}

\newcommand{\LowNetNoiseTable}{%
\begin{table*}[t]\centering\scriptsize\setlength{\tabcolsep}{3.5pt}
\caption{Clean-trained fixed-weight reconstruction with Gaussian image noise before complete Laplacian encoding. No input/output clipping or clean mean is used. Positive noise levels have three paired fixed realizations; $\pm$ is three-fit sample SD after realization/patch averaging. $\pm{<}0.001$ denotes a nonzero SD below 0.001 dB.}\label{tab:noise}
\begin{tabular}{lrrrrr}\toprule
Model & $P$ & $\sigma_{255}$ & $R=1$ & $R=2$ & $R=3$ \\
\midrule
baseline177 & 177 & 0 & $43.322\pm0.001$ & $61.958\pm0.005$ & $76.585\pm0.016$ \\
baseline177 & 177 & 5 & $33.637\pm0.001$ & $34.284\pm{<}0.001$ & $34.165\pm{<}0.001$ \\
baseline177 & 177 & 10 & $28.060\pm0.001$ & $28.271\pm{<}0.001$ & $28.144\pm{<}0.001$ \\
baseline177 & 177 & 20 & $22.160\pm0.001$ & $22.253\pm{<}0.001$ & $22.124\pm{<}0.001$ \\
unsupervised24 & 24 & 0 & $47.272\pm{<}0.001$ & $74.136\pm0.008$ & $100.741\pm0.017$ \\
unsupervised24 & 24 & 5 & $33.900\pm{<}0.001$ & $34.156\pm{<}0.001$ & $34.151\pm{<}0.001$ \\
unsupervised24 & 24 & 10 & $28.056\pm{<}0.001$ & $28.136\pm{<}0.001$ & $28.131\pm{<}0.001$ \\
unsupervised24 & 24 & 20 & $22.081\pm{<}0.001$ & $22.115\pm{<}0.001$ & $22.110\pm{<}0.001$ \\
unsupervised48 & 48 & 0 & $54.359\pm0.029$ & $85.955\pm0.036$ & $112.283\pm0.099$ \\
unsupervised48 & 48 & 5 & $34.071\pm{<}0.001$ & $34.157\pm{<}0.001$ & $34.151\pm{<}0.001$ \\
unsupervised48 & 48 & 10 & $28.088\pm0.001$ & $28.136\pm{<}0.001$ & $28.131\pm{<}0.001$ \\
unsupervised48 & 48 & 20 & $22.076\pm0.001$ & $22.116\pm{<}0.001$ & $22.110\pm{<}0.001$ \\
unsupervised72 & 72 & 0 & $56.511\pm0.003$ & $87.911\pm0.037$ & $113.067\pm0.083$ \\
unsupervised72 & 72 & 5 & $34.084\pm{<}0.001$ & $34.157\pm{<}0.001$ & $34.151\pm{<}0.001$ \\
unsupervised72 & 72 & 10 & $28.087\pm{<}0.001$ & $28.136\pm{<}0.001$ & $28.131\pm{<}0.001$ \\
unsupervised72 & 72 & 20 & $22.072\pm{<}0.001$ & $22.116\pm{<}0.001$ & $22.110\pm{<}0.001$ \\
\bottomrule\end{tabular}\end{table*}%
}

\newcommand{\LowNetRadialSupportTable}{%
\begin{table*}[t]\centering\scriptsize\setlength{\tabcolsep}{3.5pt}
\caption{Supervised single-output 100-epoch controls: free dense kernels and shell-tied kernels at mixed versus full $5\times5$ supports. These earlier metrics were scored with FP32 error arithmetic; they are not pooled with the FP64-scored core results. SDs below 0.001 dB are shown as $\pm{<}0.001$.}\label{tab:radialsupport}
\begin{tabular}{lrrrrr}\toprule
Kernels & $P$ & H/G/K/Gc & $R=1$ & $R=2$ & $R=3$ \\
\midrule
free68 & 68 & $5/3/5/3$ & $43.570\pm0.001$ & $62.305\pm0.005$ & $77.270\pm0.006$ \\
radial18 & 18 & $5/3/5/3$ & $43.501\pm0.001$ & $62.176\pm0.002$ & $77.214\pm0.007$ \\
free100 & 100 & $5/5/5/5$ & $47.496\pm{<}0.001$ & $74.289\pm0.005$ & $101.023\pm0.023$ \\
radial24 & 24 & $5/5/5/5$ & $47.273\pm{<}0.001$ & $74.123\pm0.008$ & $100.727\pm0.017$ \\
\bottomrule\end{tabular}\end{table*}%
}

\newcommand{\LowNetTiedTable}{%
\begin{table*}[t]\centering\scriptsize\setlength{\tabcolsep}{3.5pt}
\caption{H=K reduces 24 to 18 parameters with identical convolution geometry. Accuracy is three-fit mean/SD; GPU times use seed 08 with 150 events per condition on the same idle GPU. Initial moment recovery and residual encoding are included. The initializer discards old K, so initial functions are not matched. SDs below 0.001 dB are shown as $\pm{<}0.001$.}\label{tab:tied}
\begin{tabular}{lrrrrrr}\toprule
Model & $R$ & $P$ & PSNR(dB) & MSE$_{255}$ & Batch1(ms) & Batch128($\mu$s/img) \\
\midrule
untied24 & 1 & 24 & $47.272\pm{<}0.001$ & 1.422 & 0.430160 & 10.152 \\
untied24 & 2 & 24 & $74.136\pm0.008$ & 0.002929 & 0.845984 & 19.944 \\
untied24 & 3 & 24 & $100.741\pm0.017$ & 6.448e-06 & 1.247232 & 30.200 \\
tied18 & 1 & 18 & $46.340\pm{<}0.001$ & 1.774 & 0.431104 & 10.020 \\
tied18 & 2 & 18 & $72.848\pm0.007$ & 0.004086 & 0.842560 & 19.944 \\
tied18 & 3 & 18 & $98.316\pm0.020$ & 1.226e-05 & 1.248256 & 30.272 \\
\bottomrule\end{tabular}\end{table*}%
}

\begin{document}
\title{LowNet: Looped Wavelet Neural Networks with 24 Learnable Parameters for Solving Poisson Equations}
\author{Yuanhao Gong\\
	Changchun Institute of Optics, Fine Mechanics and Physics, Chinese Academy of Sciences
	\thanks{Yuanhao Gong is with the Changchun Institute of Optics, Fine Mechanics and Physics, Chinese Academy of Sciences, Changchun, China (e-mail: gong.ai@qq.com).}}
\maketitle
\begin{abstract}
Poisson equations frequently appear in various fields, such as mathematics, physics and informatics. Solving Poisson equation, however, is challenging. Traditional methods are not data-specific, although have high certainty. They require a large number of iterations due to the CFL numerical stability. In contrast, deep learning methods are data-driven and more efficient, but require accurate training data and suffer from heavy computation. To tackle these issues, we propose a wavelet-inspired Poisson solver with 24 trainable scalars. The proposed network has several advantages. First, it is compact. It has four independent $5\times5$ kernels, each represented by six parameters, leading to 24 scalars in total. Such compact format can be deployed on most devices. Second, thanks to the loop design, it achieves higher accuracy than previous method. A residual loop reuses the wavelet block without increasing its parameters. Numerical experiments confirm that such loop indeed improves the solving accuracy. On 10000 images, one, two, and three loops of the wavelet block lead to PSNRs of $47.27$, $74.14$, and $100.74$ dB, respectively. Third, the proposed method is computationally fast. Thanks to the wavelet transform, the proposed method has linear computation complexity, leading to high computation performance. In short, we propose a novel looped wavelet neural network with compactness, accuracy and efficiency for solving Poisson equations. Our method can be applied in various fields, such as computational fluid dynamics, image reconstruction, biomedical engineering, electromagnetic field simulation.
\end{abstract}
\section{Introduction}
\IEEEPARstart{P}{oisson} equations frequently appear in various fields, such as mathematics, physics and informatics. In physics, these partial differential equations are indispensable for describing potential fields, most notably in electrostatics and gravitation, where they relate the spatial distribution of a potential to its underlying source density, such as electric charge or mass. In the realm of informatics and computer vision, the Poisson equation is employed for seamless image cloning and editing, allowing for the blending of textures while preserving the gradients of the original image to ensure a natural appearance~\cite{Perez2003Poisson,Fattal2002GradientHDR}. Whether modeling the steady-state temperature distribution in a heat-conducting medium or solving complex problems in fluid dynamics, the equation provides a critical mathematical framework for understanding how local sources influence the global behavior of a system.

In computer vision, gradient-domain image processing usually leads to a Poisson equation. This occurs because the central goal of such techniques is to reconstruct a final image from a specified gradient field, rather than manipulating pixel intensities directly. To find an image whose gradients match a target field, one must solve an optimization problem that minimizes the difference between the actual image gradient and the desired gradient. Mathematically, this minimization process results in a Poisson equation. This framework is foundational to many high-end image editing tasks, most notably seamless cloning and image blending, where it allows a source patch to be integrated into a destination background by smoothly interpolating the color values while preserving the structural details of the gradients.

\begin{figure}[!tb]
	\centering
\includegraphics[width=\linewidth]{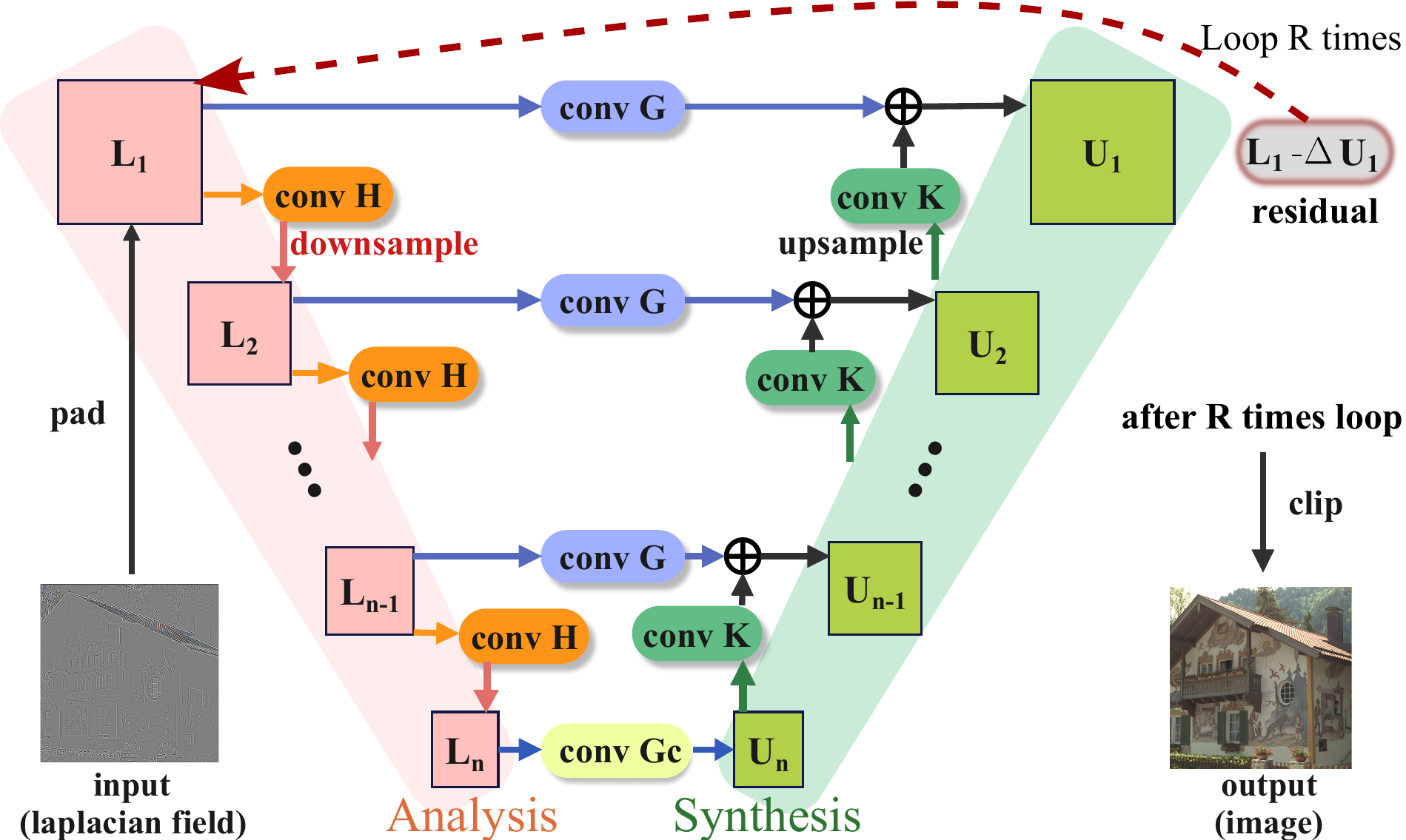}
\caption{The learnable kernels $H$, $G$, $K$ are shared across color channels and scales. The proposed network is compact. There are 24 scalar parameters in total. Moreover, the wavelet is looped to improve the accuracy. Finally, this network has linear computation complexity.}
\end{figure}

\subsection{Poisson Equation}
Gradient-domain imaging reconstructs intensities from modified derivatives through
a boundary-value problem. More specifically, these methods usually solve the following model
\begin{equation}
{\cal J}(U)=	\int_{\vec{x}\in\Omega}\frac{1}{2}|\nabla U(\vec{x}) -\vec{G}(\vec{x})|^2\mathrm{d}\vec{x}\,,
\end{equation} where $\vec{x}$ is the spatial coordinate, $\Omega$ is the imaging domain, $\vec{G}$ is the target gradient field, and $U$ is the unknown image.
Minimizing this equation leads to a Poisson equation
\begin{equation}
	\frac{\partial {\cal J}}{\partial U}=\Delta U -\nabla\cdot\vec{G}=0\,,
\end{equation}
which is 
\begin{equation}
	\Delta U =\nabla\cdot\vec{G}\equiv L\,.
\end{equation}
The $L$ is usually called source term or right hand side (RHS).
Representative applications include high-dynamic-range
compression~\cite{Fattal2002GradientHDR} and Poisson image
editing~\cite{Perez2003Poisson}. 

\subsection{Poisson Solvers}
In general, Poisson solvers can be classified into traditional solvers and learning-based solvers. Both have their advantages and disadvantages.
\subsubsection{Classical Solvers}
Fourier-based direct solution of discrete Poisson
problems has a long history~\cite{Hockney1965}, while stationary iterations and
multigrid provide established alternatives~\cite{Saad2003Iterative,
	Brandt1977Multilevel,Briggs2000Multigrid,Hackbusch1985Multigrid}.
Convolution pyramids approximate large-support linear filters with small filters
applied across a hierarchy~\cite{Farbman2011Pyramids}. 

\subsubsection{Learned Iterations and Neural PDE Solvers}
Learning a finite number of algorithmic iterations is established in sparse
coding~\cite{Gregor2010LISTA} and the broader algorithm-unrolling
literature~\cite{Monga2021Unrolling}. Learned PDE components include multigrid
prolongation~\cite{Greenfeld2019Multigrid} and convolutional pressure projection
for fluid simulation~\cite{Tompson2017Fluid}. Hsieh \emph{et al.} develop a
learned modification of iterative solvers with convergence guarantees under their
stated assumptions~\cite{Hsieh2019PDE}. These guarantees do not automatically
extend to unconstrained learned kernels. 

Physics-informed residual objectives~\cite{Raissi2019PINN} and variational neural
methods~\cite{E2018DeepRitz} provide routes to training without solution labels.
Our spectral residual objective uses the exact eigenvalues of the discrete
Dirichlet Laplacian. In exact arithmetic it equals pixel error relative to the
unique discrete solution, so it should be understood as an exactly preconditioned
objective with transform overhead, not as a new denoising regularizer or a
computationally free replacement for a direct spectral solve.

\subsection{Motivation and Contributions}
Our direct starting point is the shared-kernel wavelet network~\cite{Gong2026SWNet}. Its noise-free configuration shares analysis,
detail, and synthesis kernels across scales, uses independent kernels for the
three color channels, and has 177 trainable coefficients. 

This paper studies a compact derived model, LowNet. We share coefficients across RGB colors as well as fine scales, retain an independent detail kernel at the coarsest scale, and constrain each of four $5\times5$ kernels to six discrete radius shells. The resulting network has 24 trainable scalars. Its V-shaped hierarchy is retained. 

A single decoder call is an approximate inverse. Applying that same inverse to the current equation residual and adding the correction produces a sequence of increasingly refined estimates when the learned iteration is effective. This reuse is established in stationary numerical iterations~\cite{Saad2003Iterative}. We investigate its behavior with an unusually small learned parameterization. Repeated calls preserve the 24-scalar count, while increasing inference work. Accordingly, accuracy is reported as a function of call count rather than as a single selected score.

The study makes three contributions. 
\begin{itemize}
	\item The proposed network has only 24 parameters and thus is compact enough for most devices;
	\item The loop strategy indeed improves the accuracy and users can control this accuracy by setting the iteration number; 
	\item It reports a systematic empirical comparison of one through twelve calls, kernel support, feature paths, analysis/synthesis tying, initialization, and Gaussian image noise, with measured GPU latency and source-bound numerical evidence.
\end{itemize}

\begin{figure*}[t]
\centering
\resizebox{\textwidth}{!}{%
\begin{tikzpicture}[
  x=0.80cm,y=0.72cm,font=\footnotesize,
  >={Stealth[length=1.8mm]},
  LNstate/.style={draw=black!65,rounded corners=1pt,fill=white,align=center,
    minimum width=0.72cm,minimum height=0.52cm,inner sep=1pt},
  LNinput/.style={draw=black!65,rounded corners=2pt,fill=black!3,align=center,
    text width=2.50cm,inner sep=4pt},
  LNnote/.style={draw=black!35,rounded corners=2pt,fill=black!2,align=center,
    text width=3.40cm,inner sep=4pt},
  LNdecoder/.style={draw=blue!65!black,thick,rounded corners=2pt,fill=blue!6,
    align=center,text width=1.14cm,minimum height=0.68cm,inner sep=3pt},
  LNresidual/.style={draw=black!65,fill=white,rounded corners=1pt,
    align=center,text width=2.03cm,minimum height=0.68cm,inner sep=3pt},
  LNsum/.style={circle,draw=black!70,fill=white,inner sep=1pt,minimum size=0.33cm},
  LNflow/.style={->,thick,draw=black!75},
  LNanalysis/.style={->,thick,draw=blue!70!black},
  LNsynthesis/.style={->,thick,draw=orange!85!black},
  LNskip/.style={->,draw=green!45!black,densely dashed},
  LNobs/.style={->,draw=black!50,densely dashed}
]
\node[anchor=west,font=\small\bfseries] at (-1.85,3.75)
  {(a) One Wavelet Block $\mathcal B_\theta$: eight stored scales, four independent radial banks};

\node[LNinput] (obs) at (0,1.62)
  {$L\in\mathbb R^{3\times h\times w}$\\
   $m\leftarrow L$ };
\node[LNinput] (post) at (17.70,1.62)
  {unpadding\\$z-\bar z+m$ (per color)};
\node[draw=black!35,rounded corners=2pt,fill=black!2,align=center,
  text width=9.30cm,inner sep=4pt] at (8.85,1.95)
  {$H,G,K,G_c$: \textbf{four independent six-vectors, 24 scalars}\\
   $\begin{smallmatrix}
    \theta_8&\theta_5&\theta_4&\theta_5&\theta_8\\
    \theta_5&\theta_2&\theta_1&\theta_2&\theta_5\\
    \theta_4&\theta_1&\theta_0&\theta_1&\theta_4\\
    \theta_5&\theta_2&\theta_1&\theta_2&\theta_5\\
    \theta_8&\theta_5&\theta_4&\theta_5&\theta_8
   \end{smallmatrix}$\quad
   $\begin{gathered}\text{Same banks for R/G/B (groups 3)}\\
    \text{No activation, bias, or color mixing}\\
    H,G,K\text{ tied across fine levels};\ G_c\text{ deepest only}
   \end{gathered}$};

\node[LNstate] (x0) at (0.00,0.00) {$X_0$};
\node[LNstate] (x1) at (1.10,-0.75) {$X_1$};
\node[LNstate] (x2) at (2.20,-1.50) {$X_2$};
\node[LNstate] (x3) at (3.30,-2.25) {$X_3$};
\node[LNstate] (x4) at (4.40,-3.00) {$X_4$};
\node[LNstate] (x5) at (5.50,-3.75) {$X_5$};
\node[LNstate] (x6) at (6.60,-4.50) {$X_6$};
\node[LNstate] (x7) at (7.70,-5.25) {$X_7$};
\node[LNstate] (v7) at (10.00,-5.25) {$Y_7$};
\node[LNstate] (v6) at (11.10,-4.50) {$Y_6$};
\node[LNstate] (v5) at (12.20,-3.75) {$Y_5$};
\node[LNstate] (v4) at (13.30,-3.00) {$Y_4$};
\node[LNstate] (v3) at (14.40,-2.25) {$Y_3$};
\node[LNstate] (v2) at (15.50,-1.50) {$Y_2$};
\node[LNstate] (v1) at (16.60,-0.75) {$Y_1$};
\node[LNstate] (v0) at (17.70,0.00) {$Y_0$};

\draw[LNflow] (obs.south)--node[right,fill=white,inner sep=1pt] {$L;$ padding} (x0.north);
\foreach \i/\j in {0/1,1/2,2/3,3/4,4/5,5/6,6/7}
  \draw[LNanalysis] (x\i)--(x\j);
\draw[->,thick,draw=red!65!black] (x7.east)--node[above,fill=white,inner sep=1pt]
  {${G_c}$} (v7.west);
\foreach \i/\j in {7/6,6/5,5/4,4/3,3/2,2/1,1/0}
  \draw[LNsynthesis] (v\i)--(v\j);
\foreach \i in {0,1,2,3,4,5,6}
  \draw[LNskip] (x\i.east)--node[above,fill=white,inner sep=1pt]
    {$\,G$} (v\i.west);
\draw[LNflow] (v0.north)--(post.south);

\node[LNnote,draw=blue!50!black] at (1.80,-4.75)
  {Every blue analysis:\\
   $\operatorname{corr}(H)$: $5\times5$, stride 2, pad 2\\
   then padding};
\node[LNnote,draw=orange!65!black] at (15.95,-4.75)
  {Every orange synthesis:\\
   $\operatorname{convT}(\operatorname{flip}K)$\\
   $5\times5$, stride 2, pad 2, output pad};

\node[anchor=west,font=\small\bfseries] at (-1.85,-7.25)
  {(b) Three loops of the wavelet block: identical $\theta$, fixed $\alpha=1$, connected training graph};
\node[LNinput,text width=1.65cm] (q0) at (-0.90,-8.45) {$q_0=( L,m)$\\$u_0=0$};
\node[LNdecoder] (b1) at (1.70,-8.45) {$\mathcal B_\theta$};
\node (u1) at (3.70,-8.45) {$u_1$};
\node[LNresidual] (q1) at (5.70,-8.45)
  {$ L-\Delta u_1$\\$m-Mu_1$};
\node[LNdecoder] (b2) at (8.40,-8.45) {$\mathcal B_\theta$};
\node[LNsum] (p2) at (10.20,-8.45) {$+$};
\node (u2) at (11.30,-8.45) {$u_2$};
\node[LNresidual] (q2) at (13.30,-8.45)
  {$ L-\Delta u_2$\\$m-Mu_2$};
\node[LNdecoder] (b3) at (16.00,-8.45) {$\mathcal B_\theta$};
\node[LNsum] (p3) at (17.80,-8.45) {$+$};
\node (u3) at (19.00,-8.45) {$u_3$};
\draw[LNflow] (q0.east)--(b1.west);
\draw[LNflow] (b1.east)--(u1.west);
\draw[LNflow] (u1.east)--(q1.west);
\draw[LNflow] (q1.east)--(b2.west);
\draw[LNflow] (b2.east)--node[above] {$\delta u_2$} (p2.west);
\draw[LNflow] (p2.east)--(u2.west);
\draw[LNflow] (u2.east)--(q2.west);
\draw[LNflow] (q2.east)--(b3.west);
\draw[LNflow] (b3.east)--node[above] {$\delta u_3$} (p3.west);
\draw[LNflow] (p3.east)--(u3.west);
\draw[LNflow] (u1.north)--++(0,0.66)-| (p2.north);
\draw[LNflow] (u2.north)--++(0,0.66)-| (p3.north);
\coordinate (sharedobs) at (0,-9.50);
\draw[LNobs] (q0.south)--(q0.south |- sharedobs)--(q1.south |- sharedobs)--(q1.south);
\draw[LNobs] (q0.south)--(q0.south |- sharedobs)--(q2.south |- sharedobs)--(q2.south);
\node[fill=white,inner sep=1pt] at (8.05,-9.50)
  {Original complete $(L,m)$ reused; residual means are framed-image means};
\node[draw=violet!60!black,rounded corners=2pt,fill=violet!4,align=center,
  text width=12.65cm,inner sep=4pt] at (8.85,-10.80)
  {$\displaystyle\mathcal J=\frac{1}{3}\sum_{r=1}^{3}
      \frac{\|\Delta u_r-L\|_F^2}{NChw}$\\
   FP64 spectral residual; all three prefixes connected; fixed step one.};
\end{tikzpicture}%
}
\caption{The wavelet block and its three-prefix training unroll.
In (a), each stored field has three independently processed colors with shared coefficients. The physical one-pixel zero frame and computational five-pixel halo are
distinct. The four radial banks remain independent, while $H/G/K$ are reused
across the seven fine levels and $G_c$ is used only at the deepest level.
In (b), every occurrence of $\mathcal B_\theta$ uses the same 24 parameters.
The operator $\Delta$ produces the complete zero-framed Laplacian and $M$ the
framed-image mean. The decoder internally uses $ L$; the physics
loss instead forms the interior Dirichlet residual.}
\label{fig:lownet24-network}
\end{figure*}

\section{LowNet: Looped Wavelet Networks}
\label{sec:lownet-method}
In this section, we show the structure of the looped wavelet networks (LowNet). We show how to impose the radial symmetry in the discrete radial kernels. 
\subsection{Discrete Poisson Equation}
We consider a unit-spaced rectangular grid with an unknown RGB image
$u\in\mathbb R^{3\times h\times w}$. Each color is processed independently
using the same learned coefficients. The observed field is the \emph{complete} five-point
Laplacian
\begin{equation}
 \begin{aligned}
 L&=\Delta u\in\mathbb R^{3\times(h+2)\times(w+2)},\\
 (\Delta z)_{ij}&=z_{i-1,j}+z_{i+1,j}\\
 &\quad+z_{i,j-1}+z_{i,j+1}-4z_{ij}.
 \end{aligned}
 \label{eq:lownet-complete-L}
\end{equation}
We use the zero boundary condition, Dirichlet boundary condition. Values beyond the zero frame are also zero when evaluating the stencil. In other words, in this paper, we solve the discrete Poisson equation
\begin{equation}
	\Delta u(i,j) = f(i,j)\,,
\end{equation} with Dirichlet boundary condition.

\subsection{Shared-kernel Wavelet Solver}
The shared-kernel wavelet network can solve the Poisson equation with 177 learnable parameters~\cite{Gong2026SWNet}. It is inspired by the wavelet transform and also the deep learning methods. We noticed that the learned convolution kernels are radial symmetric. Although this property is preferred for keeping the rotation invariant property, such symmetry is not imposed in the shared-kernel wavelet network. Inspired by such radial symmetry, we propose to use radial kernels that can further reduce the number of learnable parameters. 
\subsection{Radial Kernels}
LowNet learns four independent $5\times5$ kernels $H$, $G$, $K$, and $G_c$. Considering the radial symmetry, we define a dense radial kernel as
\begin{equation}
	\begin{bmatrix}
		\theta_8&\theta_5&\theta_4&\theta_5&\theta_8\\
		\theta_5&\theta_2&\theta_1&\theta_2&\theta_5\\
		\theta_4&\theta_1&\theta_0&\theta_1&\theta_4\\
		\theta_5&\theta_2&\theta_1&\theta_2&\theta_5\\
		\theta_8&\theta_5&\theta_4&\theta_5&\theta_8
	\end{bmatrix}.
	\label{eq:lownet-shell-matrix}
\end{equation}
For integer offsets $(i,j)\in\{-2,\ldots,2\}^2$, each kernel $W$ is expanded from six unrestricted signed coefficients:
\begin{equation}
 W_{ij}=\theta^W_{i^2+j^2},\qquad
 \mathcal R=\{0,1,2,4,5,8\}.
 \label{eq:lownet-shells}
\end{equation}
The shell multiplicities are $(1,4,4,4,8,4)$, giving $4\times6=24$ trainable
scalars.  
The default initializer shell-averages the published code's distinct $H$
and $K$ kernels~\cite{Gong2026SWNet}. Its $3\times3$ $G$ is centrally zero-embedded in $5\times5$
for both $G$ and $G_c$. In the order $\mathcal R$, the nominal coefficients
are
\begin{align}
 \theta^H&=(.49,.35,.25,.105,.075,.0225),\nonumber\\
 \theta^K&=(.34,.33,.2525,.19,.0875,-.02),\nonumber\\
 \theta^G=\theta^{G_c}&=(.2992,.0957,.0306,0,0,0).
 \label{eq:lownet-initial-shells}
\end{align}
These are starting values, not algebraic constraints: $G$ and $G_c$ can
separate, their initially zero outer shells can become nonzero, and $H$
need not equal $K$. 
There are no biases, nonlinear activations, learned readout coefficients,
or cross-color mixing. The same bank is replicated across the three RGB
depthwise groups. $H$, $G$, and $K$ are shared across all scales;
$G_c$ is independent and used only at the deepest level.
\subsection{Shared-kernel Wavelet Block}
Let $P_5$ add a five-pixel zero halo and $C_5$ remove it. This computational
halo is distinct from the one-pixel physical zero frame. Denote by $D_H$ a
$5\times5$ cross-correlation with stride two and padding two, and by $C_W$
a stride-one cross-correlation with padding two. Starting from the negative
observed Laplacian, the analysis hierarchy is
\begin{equation}
 \begin{aligned}
 L_0&=P_5( L),\\
 L_{\ell+1}&=P_5D_HL_\ell,\quad 0\leq\ell<d,\\
 d&=\lceil\log_2\max(h_f,w_f)\rceil-1.
 \end{aligned}
 \label{eq:lownet-analysis}
\end{equation}

Let $U_K(z;n)$ be the transpose convolution with spatially flipped $K$,
stride two, padding two, and the output size $n$ of the matching fine field.
Its output-padding coordinate is $o=n-(2s-1)\in\{0,1\}$ for an input of size
$s$; the two axes are handled independently. The synthesis is
\begin{align}
 Y_d&=C_{G_c}X_d,\nonumber\\
 Y_\ell&=U_K(Y_{\ell+1};\operatorname{size}X_\ell)+C_GX_\ell.
 \label{eq:lownet-synthesis}
\end{align}
The recursion uses $\ell=d-1,\ldots,0$. The crop precedes each transpose
convolution; the fine-level $G$ contribution is then added at the restored
fine size. $H$ and $K$ are independent parameters. The explicit flip records
the implementation convention, even though a real radial kernel is invariant
under a $180^\circ$ spatial flip.

After the final halo crop, let $z=Y_0$ have the complete-field size.
The decoder restores its complete-field mean before removing the physical
frame:
\begin{equation}
 \mathcal B_\theta( L,m)
 =C_1\left[z-\frac{1}{h_fw_f}\sum_{i,j}z_{ij}+m\right].
 \label{eq:lownet-decoder}
\end{equation}
Mean subtraction and addition are per color. They occur before $C_1$;
they do not impose an additional exact mean projection on the delivered
cropped image. For fixed image dimensions and parameters, this entire
decoder is a linear map of $(L,m)$.
Figure~\ref{fig:lownet24-network} shows the hierarchy and its repeated use.

\subsection{Looped Wavelet Networks}
Starting from $u_0=0$,
all calls use the same 24 parameters with a fixed correction multiplier of
one:
\begin{align}
 u_1&=\mathcal B_\theta(L,m),\nonumber\\
 u_{r+1}&=u_r+
 \mathcal B_\theta\bigl(L-\Delta u_r,\;m-Mu_r\bigr).
 \label{eq:lownet-unrolling}
\end{align}
The residual retains the complete Laplacian ring. Its mean argument is the
\emph{framed-image mean residual} $m-Mu_r$, not the spatial mean of the
Laplacian residual. Neither the residual field nor the predicted image is
centered or clipped. Repeated calls add computation but no parameters.

\subsection{RHS-only spectral training}
For a predicted prefix $u_r$, the interior physics residual is
$r_r=\Delta u_r-L$. The orthonormal two-dimensional
DST-I, denoted $Q$, diagonalizes the zero-ghost Dirichlet matrix. Its
strictly positive unit-lattice eigenvalues are
\begin{equation}
 \lambda_{pq}=4\sin^2\!\frac{\pi p}{2(h+1)}
             +4\sin^2\!\frac{\pi q}{2(w+1)},
 \label{eq:lownet-eigenvalues}
\end{equation}
Here $p=1,\ldots,h$ and $q=1,\ldots,w$.
The implemented transform uses an odd FFT extension and the orthonormal
sine normalization; it is not a periodic FFT or a Neumann/DCT approximation.
For a batch of $N$ RGB observations and three connected prefixes, the
training objective is
\begin{equation}
 \mathcal J(\theta)=\frac{1}{3NChw}
   \sum_{r=1}^{3}\sum_{n=1}^{N}\sum_{c=1}^{C}
   \left\|\frac{Q r_r^{(n,c)}}{\lambda}\right\|_F^2.
 \label{eq:lownet-physics-loss}
\end{equation}
The division is elementwise and $C=3$. All three calls remain in the
computational graph; no prefix is detached.
There is no target-image term, oracle mean, relative right-hand-side
normalization, mean penalty, or outer-ring penalty in this objective.
Residual formation, eigenvalues, FFTs, and loss reductions use FP64, while
the observations, convolutional decoder, and residual encoding use FP32.
Source images are used only to construct the fixed observations. The
checkpoint is the first minimum of the full validation physics objective
over epochs $0$--$100$, including the initial model; clean-image scores do
not enter this selection.

\section{Analysis}
\label{sec:lownet-analysis-scope}

\subsection{Preconditioned Residual Equals Solution MSE}
For any interior right-hand side $b$, let $u^\star=A^{-1}b$. In exact
arithmetic, each prefix term in~\eqref{eq:lownet-physics-loss} equals its
per-pixel squared error to $u^\star$:
\begin{equation}
 \begin{aligned}
 \frac{1}{Chw}\left\|\frac{Q(Au-b)}{\lambda}\right\|_F^2
 &=\frac{1}{Chw}\|A^{-1}(Au-b)\|_F^2\\
 &=\frac{1}{Chw}\|u-u^\star\|_F^2.
 \end{aligned}
 \label{eq:lownet-loss-identity}
\end{equation}
\emph{Proof.} Write $A=Q^{\mathsf T}\operatorname{diag}(\lambda)Q$ and use
orthonormality of $Q$. Then $A^{-1}(Au-b)=u-u^\star$.\hfill$\square$

If $ L=\Delta x$ is compatible, then $u^\star=x$. For an inconsistent
outer ring, $A^{-1}b$ is still well defined, but its complete observation
need not equal the supplied $\mathcal L$. The loss does not independently
enforce that ring or the decoder's mean input. FP32 observation rounding
also distinguishes the implemented $A^{-1}b$ from the ideal original image.
The identity concerns exact arithmetic, not bit-exact agreement of an FP32
network with an exact solver. Moreover, the spectral loss costs
$O(hw\log(hw))$ per prefix and uses the same known inverse spectrum as a
direct DST solver. It is a physics preconditioner for training, not a new
inverse method or a cheap replacement for spectral inversion. In contrast,
an unpreconditioned squared residual weights the image-error sine modes
by $\lambda_{pq}^2$, underweighting the difficult low-frequency modes.

\subsection{Conditional Residual-correction Convergence}
Fix a grid and real parameters. Set $T_\theta=\mathcal B_\theta E$, acting on
interior images. For compatible data $Ex$, the exact-arithmetic errors
$e_r=x-u_r$ from~\eqref{eq:lownet-unrolling} satisfy
\begin{equation}
 e_r=(I-T_\theta)^r x.
 \label{eq:lownet-error-recurrence}
\end{equation}
\emph{Proof.} Linearity gives
$\mathcal B_\theta(Ex-Eu_r)=T_\theta e_r$, so
$e_{r+1}=(I-T_\theta)e_r$ and $e_0=x$.\hfill$\square$

On this finite-dimensional space, $\rho(I-T_\theta)<1$ is sufficient for
convergence for every image as $r\to\infty$. The stronger bound
$\|I-T_\theta\|_2\leq q<1$ gives
$\|e_r\|_2\leq q^r\|x\|_2$. Neither condition is imposed by the three-prefix
training loss or established by our finite test curves. Even a convergent
nonnormal operator need not yield monotonically decreasing Euclidean error
at every finite prefix. The recurrence is an exact-arithmetic statement;
FP32 residual encoding and updates can depart from it near a numerical
plateau. Additional inference calls therefore require measurement rather
than an unconditional convergence or monotonic-improvement claim.

\subsection{Inversion does not denoise}
For the evaluated image-domain observation model $y=x+\sigma z$, with
independent unit-variance Gaussian entries in $z$, the observed field is
$\Delta y$ and the recovered mean is $My$. Exact Dirichlet inversion recovers
$y$, including its noise. It does not recover the unknown clean $x$.
Writing $S_r=I-(I-T_\theta)^r$ and $n=Chw$, the linear-model risk conditional
on $x$ is
\begin{equation}
 \frac{1}{n}\mathbb E\|S_ry-x\|_2^2
 =\frac{1}{n}\|(S_r-I)x\|_2^2
  +\frac{\sigma^2}{n}\|S_r\|_F^2.
 \label{eq:lownet-noise-risk}
\end{equation}
Thus exact inversion, $S_r=I$, has normalized per-pixel noise MSE
$\sigma^2$. With $\sigma=\sigma_{255}/255$, this is $\sigma_{255}^2$ in
255-squared units. An approximate solver or finite prefix can trade
inversion bias against noise gain; smaller equation residual does not
guarantee smaller error to the clean image. This analysis concerns Gaussian
noise added \emph{before} Laplacian encoding, without clipping. Independent
noise on Laplacian samples is a different, potentially incompatible
observation model and was not a substitute for that comparison.

\subsection{Kernel Scale Ambiguity}
For any real $a\ne0$, replacing $H$ by $aH$ and $K$ by $K/a$, while leaving
$G$ and $G_c$ unchanged, preserves the exact decoder on every allowed input.
The analysis fields become $X'_\ell=a^\ell X_\ell$. At the
deepest scale $V'_d=a^dV_d$. Induction in~\eqref{eq:lownet-synthesis} gives
$V'_\ell=a^\ell V_\ell$, because synthesis contributes one factor $1/a$.
Consequently $V'_0=V_0$; the final mean restoration and crops are unchanged.
Repeated whole-decoder calls are unchanged.

Unique Dirichlet solutions therefore do not imply unique learned kernels.
Post-training reciprocal scaling can aid kernel comparison, but it does
not prove complete parameter identifiability and need not preserve an
Adam optimization trajectory or FP32 bit patterns. Within the constrained
$H=K$ variant, this particular gauge reduces to $a=\pm1$ for nonzero $H$;
the discrete sign ambiguity remains. No general uniqueness theorem for
$G$ versus $G_c$, or for degenerate or weakly excited paths, is asserted.

\subsection{No Orthogonal Guarantee}
The six-shell construction gives exact $D_4$ symmetry of each discrete
$5\times5$ kernel. It does not imply arbitrary-angle isotropy or exact
$D_4$ equivariance of the complete finite decoder: downsampling phase,
odd/even grids, halos, and crops also enter the operator. Similarly, tying
real radial $H=K$ makes the corresponding single-scale transpose convolution
the adjoint of analysis under the implemented padding convention. It does
not make the complete V hierarchy an orthogonal or perfect-reconstruction
wavelet transform. Such a transform requires an appropriate complete
filter bank, including complementary detail bands and boundary treatment;
kernel equality alone supplies neither condition.

\section{Experiments}
\label{sec:experiments}
\subsection{Data, Partition, and Observation Construction}
We use 100,000 uint8 RGB patches of size $128\times128$. Its SHA-256 identity, axis conversion, cached indices, and evaluation arrays are recorded in the accompanying evidence package. We normalize intensities by 255. Each patch is encoded by the fixed five-point Laplacian after appending the one-pixel zero frame in~\eqref{eq:lownet-complete-L}. 

The fixed partition contains 79,992 training, 10,000 validation, and 10,008 test patches. Exact byte-equivalent rotations and reflections are assigned as whole $D_4$ groups, giving 9,997/1,250/1,250 groups, respectively. This prevents exact augmented clones from crossing partitions. It cannot establish separation of scenes, original images, or overlapping patches, because those identifiers were not supplied. The 10,008 patches should therefore not be treated as 10,008 independent scenes. Earlier design rounds have used this test partition; the reported extensions retain frozen checkpoints and all specified calls, without representing an untouched final external test.

\subsection{Training and Reference Configurations}
The primary model uses the deterministic projected initializer in~\eqref{eq:lownet-initial-shells}. Each of three fits uses Adam~\cite{Kingma2015Adam} with learning rate $10^{-5}$, batch size 128, and 100 epochs. The seeds 20261008, 20261009, and 20261010 change the training random/shuffle stream while keeping the initial kernel values fixed. These fits measure repeatability around a common initialization; random kernel initialization is examined separately. The three-prefix loss~\eqref{eq:lownet-physics-loss} trains connected calls with fixed correction multiplier one. The first minimum of the complete validation physics loss over epochs 0--100 selects the checkpoint, including the untrained epoch-zero candidate.

The adapted Original177 reference has three color-specific sets of dense $H_5/G_3/K_5$ kernels, shared across scales, with $3(25+9+25)=177$ coefficients. We train it for 100 epochs using single-call target-image MSE and select the first full-validation MSE minimum. Its multiscale mean and observation conventions match our evaluation. Calls after the first are an external residual-loop extension evaluated here. This is a controlled adaptation of the published clean architecture, not a replication of the original paper's 2,000-epoch, separate-dataset or approximately 21K-parameter noisy experiments~\cite{Gong2026SWNet}.

A supervised radial24 reference uses the same 24-scalar architecture, single-call image MSE, and the corresponding single-call validation selector. The RHS-only and supervised comparisons consequently change training prefixes, loss arithmetic, and validation objective together. Their differences cannot isolate a causal benefit of removing labels. The exact-arithmetic loss identity nevertheless explains why similar performance is plausible.

Two independent-path variants repeat the scalar input into $C_f=2$ or 3 depthwise feature paths and average their outputs with a fixed readout. Each path has four six-shell banks, so their counts are $24C_f=48$ or 72. There is no dense feature or color mixing. Their initial $H/K$ copies match the scalar path; deterministic balanced perturbations of $G/G_c$ preserve the average initial external function while allowing branch gradients to differ. They use the same three-prefix physics training protocol. Dense $C_f\times C_f$ feature-width experiments belong to a different supervised control family and are identified separately.

The tied18 variant keeps every support $5\times5$, shares the six coefficients of $H$ and $K$, and retains independent $G$ and $G_c$. It has 18 coefficients and uses the primary three-prefix training protocol. Its shared analysis/synthesis bank starts from the projected $H$; the former $K$ initializer is discarded. Thus both the initial operator and optimization geometry change. A second 18-parameter variant, termed mixed radial18, instead keeps independent $H/K$ with $5\times5$ support and $G/G_c$ with $3\times3$ support. The two 18-parameter models must be distinguished.

\subsection{Image Noise and Error Metrics}
The noise experiment first forms $y=x+\sigma_{255}z/255$, then computes the complete Laplacian of $y$. We use $\sigma_{255}=5,10,20$, with independent unit Gaussian image pixels and no clipping. Three fixed per-index streams, seeded 20261011--20261013, are reused across models, fits, positive noise levels, and calls. This creates paired comparisons and spatially correlated Laplacian noise. All models are clean-trained and frozen; none is fitted on these noisy observations. The target for the noise tables is the original clean image. The mean input is recovered from the noisy observed field.

For each image, squared errors are averaged over all RGB pixels. We report $\MSE=255^2\|u-x\|_F^2/(3hw)$ and per-image PSNR $10\log_{10}(255^2/\MSE)$. Test PSNR is the arithmetic mean of these per-image values; it is not PSNR computed from the aggregate MSE. Noise metrics first average the fixed realizations within each fit. Unless stated otherwise, a table entry is the mean of the three fit-level test means, with sample standard deviation across those fits. This SD is neither a standard error nor a scene-level confidence interval. FP64 error arithmetic is used for the main clean, noise, tied, and random-initialization experiments. Earlier kernel-support/dense-width controls retain their original FP32 metric arithmetic and are not pooled into the primary results.

\subsection{Timing and Reproducibility}
The implementation uses PyTorch~\cite{Paszke2019PyTorch}. Convolutions and residual encoding use FP32; TF32 and automatic mixed precision are disabled. Kernel expansion is cached before inference. Controlled tied18/24 timing uses an otherwise idle NVIDIA RTX 6000 Ada GPU, resident complete observations, and a representative seed-20261008 checkpoint. Each condition receives 20 warm-up calls and 50 CUDA-event samples in each of three randomly ordered rounds. We report the pooled median of 150 event samples, for batch sizes one and 128. Timing includes initial FP64 moment recovery, decoder calls, and intermediate residual encoding and mean calculation. It excludes source-image encoding, disk I/O, host-to-device transfer, checkpoint loading, and kernel-cache preparation.

Other size/width measurements report their original median of round medians and are interpreted within their own experiment. Training wall times can include other users' GPU loads, so epoch counts are the controlled optimization budget and training minutes are not used to rank architectural cost. Larger $2048\times2048$ inputs were tested for timing only. No accuracy claim at that resolution follows from those measurements. The evidence package records source/checkpoint hashes, ordered test identities, training histories, full numerical tables, and figure-generation code; it also distinguishes observed metrics from theoretical operation counts.

\section{Results and Ablations}
\label{sec:results}
\subsection{Clean Reconstruction and Increasing Call Count}
\begin{figure*}[t]
\centering\includegraphics[width=\textwidth]{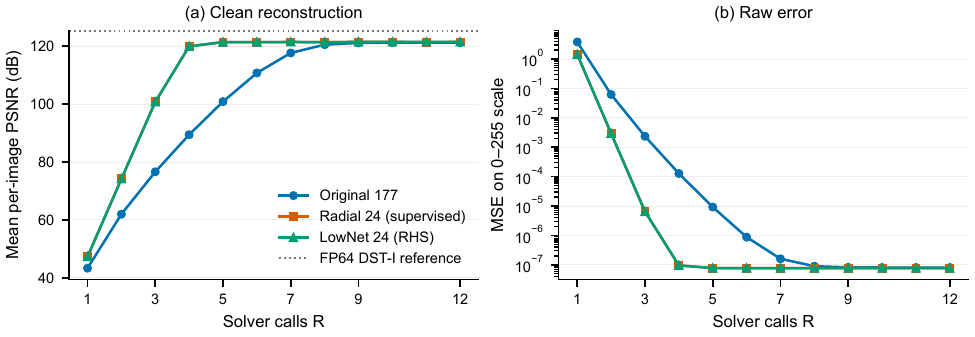}
\caption{Frozen-weight clean reconstruction for all measured call counts. Curves use mean per-image PSNR and mean pixel MSE on the same ordered 10,008 test patches. The exact CPU FP64 DST-I reference uses the same FP32 complete Laplacian and is a numerical accuracy reference; a matched GPU runtime was not measured. Fit SD is small relative to the plotted range.}
\label{fig:clean-loops}
\end{figure*}

Table~\ref{tab:clean} and Fig.~\ref{fig:clean-loops} report every measured prefix from one through twelve using frozen weights. Original177 improves from 43.322 dB at one call to 61.958 and 76.585 dB at two and three. LowNet24 reaches 47.272, 74.136, and 100.741 dB. The differences at these prefixes are 3.950, 12.178, and 24.157 dB. Its corresponding $\MSE$ values are 1.421743, $2.928823\times10^{-3}$, and $6.448140\times10^{-6}$, compared with 3.776137, $6.158993\times10^{-2}$, and $2.332296\times10^{-3}$ for Original177. The 24-parameter configuration therefore supplies a substantially more effective finite-call inverse on these patches, while reducing the learned scalar count by 86.4\%. This is a comparison of complete configurations, which includes support, coarse-path, color sharing, radial coordinates, and training differences.

\LowNetCleanTable

At four calls, LowNet24 reaches 119.924 dB and $\MSE=9.512810\times10^{-8}$. Five calls yield 121.378 dB, after which scores remain near 121.4 dB with small nonmonotone fluctuations. The measured fifth-to-twelfth-call behavior is a numerical plateau, rather than an extrapolated convergence curve. Original177 approaches approximately 121.2 dB by calls ten through twelve. Additional LowNet calls thus provide substantial benefit through four or five calls on this grid, followed by little accuracy return for their extra work. This empirical pattern does not impose a universal call-count stopping rule on other grids or observations.

The saved exact solver reference applies CPU FP64 SciPy DST-I inversion to the same FP32 Laplacian input over the complete test set. It yields mean per-image PSNR 125.171 dB and $\MSE=4.817291\times10^{-8}$. The learned plateau remains less accurate. Finite-precision Laplacian observations also explain why this direct solve need not reproduce the original uint8-derived image bit for bit. These data support a convolutional approximate inverse, without establishing accuracy superiority over direct inversion or a GPU speed advantage over it.

The supervised radial24 scores are 47.273, 74.123, and 100.727 dB for the first three calls. Relative to them, the RHS-only model changes PSNR by $-0.000326$, $+0.012870$, and $+0.014515$ dB. These are small differences compared with the architectural gain over Original177. At higher calls the two radial curves reach essentially the same plateau. The observed equivalence is consistent with Proposition~1; it does not establish that one objective universally optimizes better, because the prefixes and checkpoint objectives differ as described in Section~\ref{sec:experiments}.

\subsection{Kernel Support and Radial Compression}
The supervised kernel-support controls in Table~\ref{tab:radialsupport} separate two useful comparisons. With $H/K$ at $5\times5$ and $G/G_c$ at $3\times3$, dense free68 and mixed radial18 reach 43.570 and 43.501 dB after one call, respectively. With all supports $5\times5$, dense free100 and radial24 reach 47.496 and 47.273 dB. Thus the shell constraint retains most of the finite-call precision of the matched-support dense models at a much smaller scalar count. Dense free100 remains slightly more accurate: its three-call score is 101.023 dB, versus 100.727 dB for the supervised radial24 reference. Compactness does not imply the best attainable accuracy within the wider dense family. Radial projection also changes the initial synthesis kernel and raw shell coordinates accumulate pixel gradients within each shell, so these matched-support comparisons do not isolate symmetry alone.

\LowNetRadialSupportTable
\begin{figure*}[t]
\centering\includegraphics[width=\textwidth]{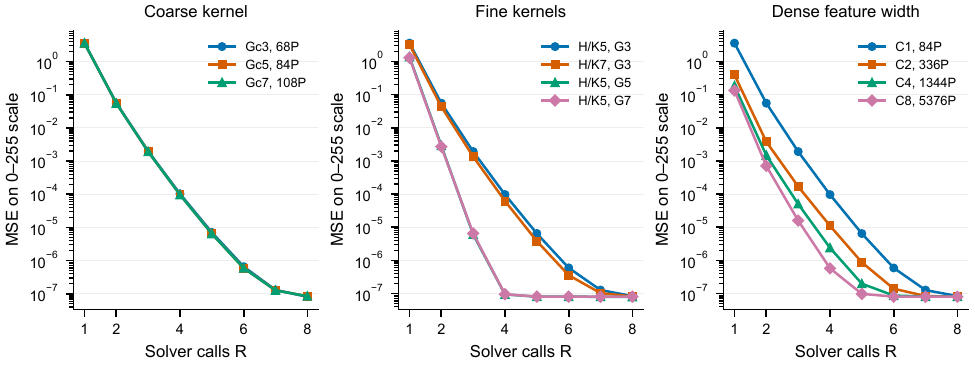}
\caption{Earlier supervised single-output kernel-size and dense-width controls. Panels vary the coarse detail support, fine-kernel supports, and dense $C_f\times C_f$ feature width. These dense-width controls differ from the independent 24/48/72-parameter paths in Table~\ref{tab:noise}; the experiment families are not pooled.}
\label{fig:support-width}
\end{figure*}

Increasing $G/G_c$ together from $3\times3$ to $5\times5$ in the radial model raises the one-call score from 43.501 to 47.273 dB and the three-call score from 77.214 to 100.727 dB. Because both detail supports change, this comparison cannot assign the gain to either kernel individually. It also changes the number of shell coordinates from 18 to 24. Earlier coarse-only experiments, detailed in the supplement, show much smaller changes when only $G_c$ is enlarged while the fine-scale detail support stays $3\times3$. The evidence favors examining the whole hierarchy instead of attributing the improvement solely to a large coarsest kernel.

\subsection{Additional Feature Paths and Dense Width}
Independent-path physics training improves early-call precision as the coefficient count grows. The 48-parameter model obtains 54.359, 85.955, and 112.283 dB at calls one through three; the 72-parameter model obtains 56.511, 87.911, and 113.067 dB. Relative to LowNet24, the respective first-call gains are 7.087 and 9.239 dB, and the three-call gains are 11.542 and 12.326 dB. The increment from 48 to 72 is smaller, particularly at three calls. These results expose a capacity--call tradeoff: a larger bank can improve a fixed small number of calls, whereas repeated scalar-path calls can approach the floating-point plateau without increasing stored trainable coefficients.

The older dense-width experiments use a different mechanism. Four dense kernels with supports $k_H,k_G,k_K,k_c$ and width $C_f$ have $C_f^2(k_H^2+k_G^2+k_K^2+k_c^2)$ total RGB-shared trainable coefficients. For $H/K=5$, $G=3$, $G_c=5$, widths 1, 2, 4, and 8 contain 84, 336, 1,344, and 5,376 coefficients and reach 43.604, 53.662, 57.224, and 58.696 dB at one call. Their theoretical convolution MAC counts rise from 2.020 to 129.270 million per native-$128$ call. The batch-one one-call latencies in that experiment rise from approximately 0.430 to 0.691 ms. Small-grid latency is affected by launch and memory costs as well as arithmetic, so MAC ratios do not predict timing ratios. Complete size, width, and timing tables appear in the supplement.

Parameter sharing and operation reduction are likewise distinct. The all-$5\times5$ scalar geometry requires approximately 3.481 million convolution MACs per native-$128$ call, compared with 2.002 million for the reference $5/3/5$ geometry. Expanding a shell kernel once for cached inference does not avoid its ordinary convolution operations. The 24 coefficients occupy 96 raw FP32 bytes, but this excludes caches, activations, framework state, and checkpoint containers. The strongest benefit established by the count is a compact learned representation.

\subsection{Tying Analysis and Synthesis: 24 versus 18}
The tied18 model removes one six-vector by enforcing $H=K$. Table~\ref{tab:tied} gives matched training-budget accuracy and controlled same-GPU timing. Its first three PSNRs are 46.340, 72.848, and 98.316 dB, lower than untied24 by 0.933, 1.289, and 2.426 dB. Its $\MSE$ is approximately 1.248, 1.395, and 1.902 times larger. The direction of the difference is consistent across the three fitted shuffle streams.

\LowNetTiedTable
The parameter reduction does not change the convolution geometry. For batch one, untied24 requires 0.430, 0.846, and 1.247 ms at calls one through three; tied18 requires 0.431, 0.843, and 1.248 ms. For batch 128, untied24 requires 10.152, 19.944, and 30.200 $\mu$s per image, and tied18 requires 10.020, 19.944, and 30.272 $\mu$s. These small alternating differences provide no consistent acceleration from tying. Since the tied initializer and accumulated $H/K$ gradients change, the experiment establishes the observed 100-epoch outcome of this constrained configuration, not a universal irreducible accuracy penalty. Within the tested setting, keeping $H$ and $K$ independent is preferable.

\subsection{Kernel Repeatability and Random Initialization}
Three fits from the same projected initializer have very similar learned shell curves. Maximum pairwise relative Frobenius differences of the expanded kernels are 0.01883\%, 0.01994\%, 0.06469\%, and 2.52326\% for $H,G,K,G_c$, respectively. Shell multiplicities weight this comparison so that it measures kernel-pixel differences rather than an unweighted six-vector distance. The coarsest kernel varies more, while delivered reconstruction remains stable. Reciprocal analysis/synthesis scale alignment changes the maxima for $H/K$ only slightly. The supplement shows all four learned curves and the explicit signed coefficients.

\begin{figure*}[t]
\centering\includegraphics[width=\textwidth]{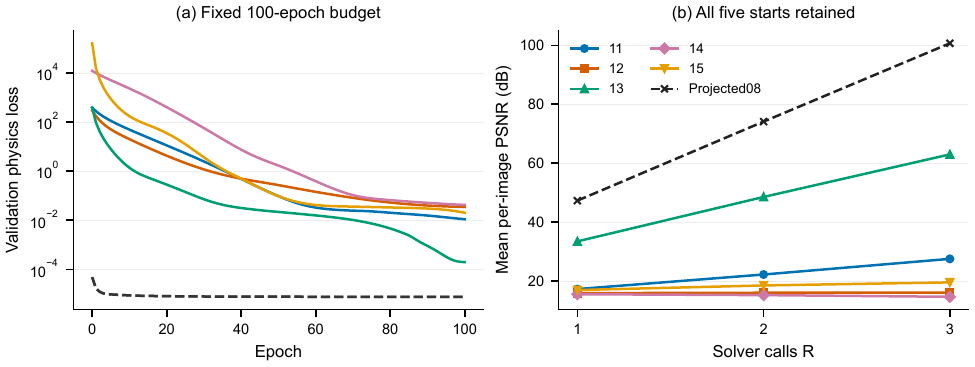}
\caption{Five completed random-initialization fits versus the preserved projected-initialization reference. All random runs use the fixed 100-epoch budget and retain their best epoch-100 checkpoints. The full spread and call dependence are shown; poor accuracy is not labeled numerical divergence, and continued validation descent limits conclusions about longer optimization.}
\label{fig:init}
\end{figure*}

Repeatability near a common initialization is distinct from robustness to arbitrary initialization. We train five additional models with all 24 coefficients independently drawn from $\mathcal N(0,0.2^2)$, a common shuffle stream, and the same 100-epoch physics-training budget. Every run completes and selects epoch 100. Their three-call PSNRs are 27.494, 16.037, 62.982, 14.607, and 19.503 dB, versus 100.722 dB for the preserved single-fit projected reference. Figure~\ref{fig:init} retains every random run. One model deteriorates as calls increase, from 15.476 dB at one call to 14.607 dB at three; finite-loop improvement is therefore not automatic for arbitrary learned coefficients.

Raw random-fit kernel differences are large: pairwise relative distances range from approximately 48\% to 195\% across the four banks. Removing the known reciprocal $H/K$ gauge still leaves substantial differences. Because these fits do not have comparable low solution error, their kernels cannot be used to demonstrate multiple equally optimal solutions. The experiment instead shows pronounced fixed-budget optimization sensitivity. All five validation losses are still decreasing near the budget endpoint; longer training or different initialization scales may change the outcome. 

\subsection{Gaussian Image Noise: Inversion versus Denoising}
\begin{figure*}[t]
\centering\includegraphics[width=\textwidth]{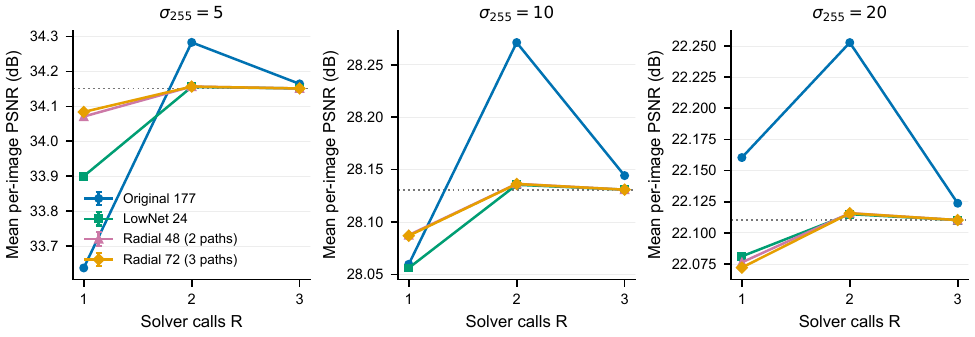}
\caption{Gaussian noise is added to RGB image pixels before the complete Laplacian is computed. Scores use the clean target, frozen clean-trained models, paired noise realizations, and recovered noisy means. Improving the equation solve approaches recovery of the noisy image. The noise-variance references describe inversion, not a claimed denoising bound for estimators with an image prior.}
\label{fig:noise}
\end{figure*}

Table~\ref{tab:noise} and Fig.~\ref{fig:noise} compare clean-trained Original177 and the 24/48/72-parameter physics models under the requested image-first noise protocol. At $\sigma_{255}=10$, LowNet24 has clean-target PSNR 28.056, 28.136, and 28.131 dB at calls one through three. Its corresponding $\MSE$ values are 101.740, 99.893, and 100.002. At three calls, $\sigma_{255}=5$ and 20 give 34.151 and 22.110 dB, with MSE near 25 and 400. These values track the noise-variance floor of exact inversion.

\LowNetNoiseTable

The additional independent paths improve clean inversion substantially, yet have nearly the same three-call noisy scores. At $\sigma_{255}=10$, 48 and 72 parameters yield 28.131 dB, essentially matching LowNet24. The measured noise-energy gain of LowNet24 at three calls is approximately 1.000008. This agrees with the near-identity limit in~\eqref{eq:lownet-noise-risk}. The slight two-call advantage reflects an incomplete inverse attenuating some noise together with solution components; it is not evidence that stronger equation accuracy improves clean-image denoising.

Original177 obtains 28.060, 28.271, and 28.144 dB at $\sigma_{255}=10$. Its two-call clean-target score can exceed that of LowNet24 even though its clean equation solve is less accurate. This illustrates the bias/noise tradeoff rather than contradicting the clean comparison. The present evidence therefore supports better Poisson reconstruction at small parameter count, but supplies no general superiority in image denoising. Achieving such a goal requires a distinct target, prior, regularization, or noise-aware training protocol. Published tests with noise independently added to Laplacian samples and a larger multichannel model~\cite{Gong2026SWNet} answer a different question and are not inserted into this table.

\section{Discussion and Limitations}
The experiments show that a useful small solver can be obtained by sharing across colors and scales, preserving an independent coarse detail bank, and using signed radial coordinates. The benefit is supported by both a matched-support dense comparison and a complete frozen call curve. The resulting 24-scalar configuration is a practical compromise within the tested family: independent extra paths improve early-call accuracy, whereas further tying to 18 coefficients loses accuracy with little timing change. A global optimality claim would require a broader search and an independent final evaluation. The present results do not establish that 24 is the smallest possible useful count.

Seven of the nine primary RHS-only fits (24/48/72 parameters, three fits each) select epoch 100. This is a fixed comparison budget rather than evidence of full optimization convergence. The four learned kernels should be interpreted as coupled components of one multiscale inverse. Their plots resemble smoothing/detail filters, but classical wavelet identities are not imposed. In particular, different analysis and synthesis kernels need not be a verified biorthogonal pair, and equal radial kernels do not establish perfect reconstruction. The scale ambiguity proves that inspecting a single kernel norm cannot identify the external inverse. The random-fit study further indicates that the convenient projected initializer carries substantial optimization value under this learning rate.

The meaning of RHS-only training also deserves precision. The optimizer and checkpoint selector need no clean solution labels once observations are prepared. However, the loss uses a known spectral inverse of the same fixed Dirichlet operator, and Proposition~1 makes its target implicit. This provides an explicit, well-conditioned training criterion for a convolutional surrogate; it does not remove the numerical knowledge needed to define the problem. For variable coefficients, irregular domains, mixed boundaries, or incomplete Laplacian observations, the diagonal DST spectrum and complete-field mean identity would need to be replaced. Those extensions are outside the evaluated evidence.

The data and computation impose additional limits. Exact $D_4$ grouping controls augmented duplicates but cannot verify source-scene independence. The repeatedly inspected test partition leaves the study exploratory. Only native $128\times128$ patches have reconstruction accuracy measurements; the multiscale topology and larger-image timing do not establish resolution-independent accuracy. Very high clean PSNR values describe a compatible synthetic observation with FP32 arithmetic, not perceptual robustness on real derivative sensors. Three fixed noise realizations quantify a paired benchmark, rather than all possible corruption processes. Future validation should reserve unseen source images, vary boundary and input precision, and benchmark the learned loop against tuned direct and iterative solvers under matched hardware.

\section{Conclusion}
We presented LowNet, a V-shaped multiscale Poisson solver with four six-shell $5\times5$ kernels shared across colors and fine scales. An independent coarsest bank and whole-decoder residual reuse retain a compact 24-scalar representation. The RHS-only spectral objective equals discrete solution MSE in exact arithmetic; the analysis also identifies conditional loop convergence, kernel scale nonuniqueness, and the limit of inversion under image-domain noise. On the supplied patch benchmark, the first three calls obtain 47.272, 74.136, and 100.741 dB, and five calls reach an approximately 121.4-dB numerical plateau. Kernel/path ablations and controlled timing favor independent analysis/synthesis banks under the tested budget, while random initializations reveal substantial optimization sensitivity. The evidence supports an accurate compact convolutional approximation in this discrete setting. The proposed method can be applied in various image processing applications~\cite{8803492,TANG2023103710,ssgong,EGGS, legs,s23084063,osbf,EDH, irs,Wei2025,11059871,Guo2026,huo2026lvslocalviewsynthesis}.

\bibliographystyle{IEEEtran}
\bibliography{references}

@misc{huo2026lvslocalviewsynthesis,
      title={LVS: Local View Synthesis from Relative Camera Pose by Reusing Previous Views}, 
      author={Qizhou Huo and Xuan Sun and Yongfei Guo and Zhipeng Wang and Yuanhao Gong},
      year={2026},
      eprint={2610.12127},
      archivePrefix={arXiv},
      primaryClass={cs.CV},
      url={https://arxiv.org/abs/2610.12127}, 
}

@ARTICLE{11059871,
  author={Gong, Yuanhao},
  journal={IEEE Access}, 
  title={Removing Scattered Light in Biomedical Images via Total Variation Guided Filter}, 
  year={2025},
  volume={13},
  number={},
  pages={114495-114505},
  doi={10.1109/ACCESS.2025.3584403}}

@Article{s23084063,
AUTHOR = {Xu, Meng and Zhang, Zhihuang and Gong, Yuanhao and Poslad, Stefan},
TITLE = {Regression-Based Camera Pose Estimation through Multi-Level Local Features and Global Features},
JOURNAL = {Sensors},
VOLUME = {23},
YEAR = {2023},
NUMBER = {8},
ARTICLE-NUMBER = {4063},
URL = {https://www.mdpi.com/1424-8220/23/8/4063},
PubMedID = {37112404},
ISSN = {1424-8220},
DOI = {10.3390/s23084063}
}

@ARTICLE{EDH,
  author={Gong, Yuanhao and Lin, Zewei},
  journal={IEEE Sensors Journal}, 
  title={EDHnet: An Event-Based Dual-Stream Hybrid Network for Image Motion Deblurring}, 
  year={2024},
  volume={24},
  number={20},
  pages={32884-32897},
  doi={10.1109/JSEN.2024.3454153}}

@ARTICLE{irs,
  author={Gong, Yuanhao},
  journal={IEEE Access}, 
  title={IRSnet: An Implicit Residual Solver and Its Unfolding Neural Network With 0.003M Parameters for Total Variation Models}, 
  year={2025},
  volume={13},
  number={},
  pages={10289-10298},
  doi={10.1109/ACCESS.2025.3528637}}

@article{TANG2023103710,
title = {Feature preserving 3D mesh denoising with a Dense Local Graph Neural Network},
journal = {Computer Vision and Image Understanding},
volume = {233},
pages = {103710},
year = {2023},
issn = {1077-3142},
doi = {https://doi.org/10.1016/j.cviu.2023.103710},
url = {https://www.sciencedirect.com/science/article/pii/S1077314223000905},
author = {Wenming Tang and Yuanhao Gong and Guoping Qiu}
}

@inproceedings{legs,
author = {Guo, Yongfei and Huo, Qizhou and Sun, Xuan and Gong, Yuanhao},
title = {LEGS: Laplacian-Enhanced Gaussian Splatting with a Nonlinear Weighted Loss},
year = {2026},
isbn = {9798400728105},
publisher = {Association for Computing Machinery},
address = {New York, NY, USA},
url = {https://doi.org/10.1145/3822516.3834868},
doi = {10.1145/3822516.3834868},
booktitle = {Proceedings of the 31st ACM International Conference on 3D Web Technology},
articleno = {23},
numpages = {5},
location = {Doha, Qatar},
series = {Web3D '26}
}

@ARTICLE{osbf,
  author={Gong, Yuanhao},
  journal={IEEE Access}, 
  title={OSBF: One-Sided Box Filter for Edge-Preserving Image Processing}, 
  year={2025},
  volume={13},
  number={},
  pages={61149-61160},
  doi={10.1109/ACCESS.2025.3555434}}

@inproceedings{EGGS,
author = {Gong, Yuanhao},
title = {EGGS: Edge Guided Gaussian Splatting for Radiance Fields},
year = {2024},
isbn = {9798400706899},
publisher = {Association for Computing Machinery},
address = {New York, NY, USA},
url = {https://doi.org/10.1145/3665318.3677148},
doi = {10.1145/3665318.3677148},
booktitle = {Proceedings of the 29th International ACM Conference on 3D Web Technology},
articleno = {5},
numpages = {5},
location = {Guimar{\~a}es, Portugal},
series = {Web3D '24}
}

@InProceedings{ssgong,
author="Gong, Yuanhao
and Wang, Qicong
and Yang, Chenhui
and Gao, Yahui
and Li, Cuihua",
editor="Jiang, Xiaoyi
and Petkov, Nicolai",
title="Symmetry Detection for Multi-object Using Local Polar Coordinate",
booktitle="Computer Analysis of Images and Patterns",
year="2009",
publisher="Springer Berlin Heidelberg",
address="Berlin, Heidelberg",
pages="277--284",
isbn="978-3-642-03767-2"
}

@INPROCEEDINGS{8803492,
  author={Gong, Yuanhao and Yin, Hui and Liu, Jingxin and Liu, Bozhi and Qiu, Guoping},
  booktitle={2019 IEEE International Conference on Image Processing (ICIP)}, 
  title={Soft Tissue Removal in X-Ray Images by Half Window Dark Channel Prior}, 
  year={2019},
  volume={},
  number={},
  pages={3576-3580},
  doi={10.1109/ICIP.2019.8803492}}

@article{Gong2026SWNet,
  author  = {Gong, Yuanhao and Tang, Tan and Liu, Qianyan},
  title   = {Shared-Kernel Wavelet Networks for Near-Sensor {Poisson} Image Reconstruction},
  journal = {IEEE Sensors Journal},
  year    = {2026},
  volume  = {26},
  number  = {14},
  pages   = {21594--21602},
  doi     = {10.1109/JSEN.2026.3702036},
  url     = {https://github.com/gyh8/swnet}
}

@article{Perez2003Poisson,
  author  = {P{\'e}rez, Patrick and Gangnet, Michel and Blake, Andrew},
  title   = {{Poisson} Image Editing},
  journal = {ACM Transactions on Graphics},
  year    = {2003},
  volume  = {22},
  number  = {3},
  pages   = {313--318},
  url     = {https://legacy.sites.fas.harvard.edu/~cs278/papers/poisson.pdf}
}

@inproceedings{Farbman2011Pyramids,
  author    = {Farbman, Zeev and Fattal, Raanan and Lischinski, Dani},
  title     = {Convolution Pyramids},
  booktitle = {Proceedings of the 2011 SIGGRAPH Asia Conference},
  year      = {2011},
  pages     = {1--8},
  publisher = {ACM},
  doi       = {10.1145/2024156.2024209},
  url       = {https://pages.cs.huji.ac.il/danix-lab/cglab/projects/convpyr/}
}

@book{Briggs2000Multigrid,
  author    = {Briggs, William L. and Henson, Van Emden and McCormick, Steve F.},
  title     = {A Multigrid Tutorial},
  edition   = {2},
  publisher = {Society for Industrial and Applied Mathematics},
  address   = {Philadelphia, PA, USA},
  year      = {2000},
  doi       = {10.1137/1.9780898719505},
  url       = {https://epubs.siam.org/doi/book/10.1137/1.9780898719505}
}

@book{Hackbusch1985Multigrid,
  author    = {Hackbusch, Wolfgang},
  title     = {Multi-Grid Methods and Applications},
  series    = {Springer Series in Computational Mathematics},
  volume    = {4},
  publisher = {Springer},
  address   = {Berlin, Heidelberg, Germany},
  year      = {1985},
  doi       = {10.1007/978-3-662-02427-0},
  url       = {https://link.springer.com/book/10.1007/978-3-662-02427-0}
}

@book{Saad2003Iterative,
  author    = {Saad, Yousef},
  title     = {Iterative Methods for Sparse Linear Systems},
  edition   = {2},
  publisher = {Society for Industrial and Applied Mathematics},
  address   = {Philadelphia, PA, USA},
  year      = {2003},
  doi       = {10.1137/1.9780898718003},
  url       = {https://epubs.siam.org/doi/book/10.1137/1.9780898718003}
}

@article{Hockney1965,
  author  = {Hockney, R. W.},
  title   = {A Fast Direct Solution of {Poisson}'s Equation Using {Fourier} Analysis},
  journal = {Journal of the ACM},
  year    = {1965},
  volume  = {12},
  number  = {1},
  pages   = {95--113}
}

@inproceedings{Gregor2010LISTA,
  author    = {Gregor, Karol and LeCun, Yann},
  title     = {Learning Fast Approximations of Sparse Coding},
  booktitle = {Proceedings of the 27th International Conference on Machine Learning},
  year      = {2010},
  pages     = {399--406},
  url       = {https://icml.cc/Conferences/2010/papers/449.pdf}
}

@article{Monga2021Unrolling,
  author  = {Monga, Vishal and Li, Yuelong and Eldar, Yonina C.},
  title   = {Algorithm Unrolling: Interpretable, Efficient Deep Learning for Signal and Image Processing},
  journal = {IEEE Signal Processing Magazine},
  year    = {2021},
  volume  = {38},
  number  = {2},
  pages   = {18--44},
  doi     = {10.1109/MSP.2020.3016905},
  url     = {https://arxiv.org/abs/1912.10557}
}

@article{Raissi2019PINN,
  author  = {Raissi, Maziar and Perdikaris, Paris and Karniadakis, George Em},
  title   = {Physics-Informed Neural Networks: A Deep Learning Framework for Solving Forward and Inverse Problems Involving Nonlinear Partial Differential Equations},
  journal = {Journal of Computational Physics},
  year    = {2019},
  volume  = {378},
  pages   = {686--707},
  doi     = {10.1016/j.jcp.2018.10.045},
  url     = {https://www.sciencedirect.com/science/article/pii/S0021999118307125}
}

@article{E2018DeepRitz,
  author  = {E, Weinan and Yu, Bing},
  title   = {The Deep {Ritz} Method: A Deep Learning-Based Numerical Algorithm for Solving Variational Problems},
  journal = {Communications in Mathematics and Statistics},
  year    = {2018},
  volume  = {6},
  number  = {1},
  pages   = {1--12},
  doi     = {10.1007/s40304-018-0127-z},
  url     = {https://link.springer.com/article/10.1007/s40304-018-0127-z}
}

@inproceedings{Greenfeld2019Multigrid,
  author    = {Greenfeld, Daniel and Galun, Meirav and Basri, Ronen and Yavneh, Irad and Kimmel, Ron},
  title     = {Learning to Optimize Multigrid {PDE} Solvers},
  booktitle = {Proceedings of the 36th International Conference on Machine Learning},
  series    = {Proceedings of Machine Learning Research},
  volume    = {97},
  year      = {2019},
  pages     = {2415--2423},
  publisher = {PMLR},
  url       = {https://proceedings.mlr.press/v97/greenfeld19a.html}
}

@inproceedings{Hsieh2019PDE,
  author    = {Hsieh, Jun-Ting and Zhao, Shengjia and Eismann, Stephan and Mirabella, Lucia and Ermon, Stefano},
  title     = {Learning Neural {PDE} Solvers With Convergence Guarantees},
  booktitle = {International Conference on Learning Representations},
  year      = {2019},
  url       = {https://openreview.net/forum?id=rklaWn0qK7}
}

@article{Brandt1977Multilevel,
  author  = {Brandt, Achi},
  title   = {Multi-Level Adaptive Solutions to Boundary-Value Problems},
  journal = {Mathematics of Computation},
  year    = {1977},
  volume  = {31},
  number  = {138},
  pages   = {333--390},
  url     = {https://www.wisdom.weizmann.ac.il/~achi/LOP16.pdf}
}

@article{Fattal2002GradientHDR,
  author  = {Fattal, Raanan and Lischinski, Dani and Werman, Michael},
  title   = {Gradient Domain High Dynamic Range Compression},
  journal = {ACM Transactions on Graphics},
  year    = {2002},
  volume  = {21},
  number  = {3},
  pages   = {249--256},
  url     = {https://pages.cs.wisc.edu/~dyer/cs534/papers/fattal-hdrc.pdf}
}

@inproceedings{Tompson2017Fluid,
  author    = {Tompson, Jonathan and Schlachter, Kristofer and Sprechmann, Pablo and Perlin, Ken},
  title     = {Accelerating Eulerian Fluid Simulation With Convolutional Networks},
  booktitle = {Proceedings of the 34th International Conference on Machine Learning},
  series    = {Proceedings of Machine Learning Research},
  volume    = {70},
  year      = {2017},
  pages     = {3424--3433},
  publisher = {PMLR},
  url       = {https://proceedings.mlr.press/v70/tompson17a.html}
}

@inproceedings{Kingma2015Adam,
  author    = {Kingma, Diederik P. and Ba, Jimmy Lei},
  title     = {{Adam}: A Method for Stochastic Optimization},
  booktitle = {International Conference on Learning Representations},
  year      = {2015},
  url       = {https://arxiv.org/abs/1412.6980}
}

@inproceedings{Paszke2019PyTorch,
  author    = {Paszke, Adam and Gross, Sam and Massa, Francisco and Lerer, Adam and Bradbury, James and Chanan, Gregory and Killeen, Trevor and Lin, Zeming and Gimelshein, Natalia and Antiga, Luca and Desmaison, Alban and Kopf, Andreas and Yang, Edward and DeVito, Zachary and Raison, Martin and Tejani, Alykhan and Chilamkurthy, Sasank and Steiner, Benoit and Fang, Lu and Bai, Junjie and Chintala, Soumith},
  title     = {{PyTorch}: An Imperative Style, High-Performance Deep Learning Library},
  booktitle = {Advances in Neural Information Processing Systems},
  volume    = {32},
  year      = {2019},
  pages     = {8024--8035},
  publisher = {Curran Associates, Inc.},
  url       = {https://proceedings.neurips.cc/paper/2019/hash/bdbca288fee7f92f2bfa9f7012727740-Abstract.html}
}

@Article{Wei2025,
  author   = {Wei, Zhili and Tang, Wenming and Gong, Yuanhao},
  journal  = {PeerJ},
  title    = {Bone-tissue decomposition of a single X-ray image via solving a Laplace equation},
  year     = {2025},
  issn     = {2167-8359},
  pages    = {e20016},
  volume   = {13},
  doi      = {10.7717/peerj.20016},
  url      = {https://doi.org/10.7717/peerj.20016},
}

@InProceedings{Guo2026,
  author       = {Yongfei Guo and Qizhou Huo and Chizhi Zhang and Xudong Niu and Yuanhao Gong},
  booktitle    = {International Conference on Optoelectronic Information and Computer Engineering (OICE 2026)},
  title        = {{Quarter-window Gaussian filter for image processing}},
  year         = {2026},
  editor       = {Yufei Ma and Yang Yue},
  organization = {International Society for Optics and Photonics},
  pages        = {1431710},
  publisher    = {SPIE},
  volume       = {14317},
  doi          = {10.1117/12.3122631},
  url          = {https://doi.org/10.1117/12.3122631},
}
\end{document}